\documentclass[journal]{IEEEtran}
\usepackage[switch]{lineno}

\usepackage{amsmath,amsfonts}

\usepackage{array}
\usepackage[caption=false,font=normalsize,labelfont=sf,textfont=sf]{subfig}
\usepackage{textcomp}
\usepackage{stfloats}
\usepackage{url}
\usepackage{verbatim}
\usepackage{graphicx}
\usepackage{cite}
\usepackage{multirow}
\usepackage{booktabs}
\usepackage[ruled]{algorithm2e}
\usepackage{makecell}

\usepackage[normalem]{ulem}
\usepackage{xcolor}

\begin{document}
\title{Mutual Distillation Learning For \\Person Re-Identification}

\author{Huiyuan~Fu,~\IEEEmembership{Member,~IEEE,}
        Kuilong Cui,
        Chuanming Wang,
        Mengshi Qi,~\IEEEmembership{Member,~IEEE,} \\
        Huadong Ma,~\IEEEmembership{Fellow,~IEEE}
\thanks{H. Fu, K. Cui, C. Wang, M. Qi and H. Ma are with Beijing Key Laboratory of Intelligent Telecommunications Software and Multimedia, Beijing University of Posts and Telecommunications, China. (\emph{Corresponding author: Mengshi Qi~(email:~qms@bupt.edu.cn)})}
}


\maketitle

\begin{abstract}
With the rapid advancements in deep learning technologies, person re-identification (ReID) has witnessed remarkable performance improvements. However, the majority of prior works have traditionally focused on solving the problem via extracting features solely from a single perspective, such as uniform partitioning, hard attention mechanisms, or semantic masks. While these approaches have demonstrated efficacy within specific contexts, they fall short in diverse situations. In this paper, we propose a novel approach, \textit{M}utual \textit{D}istillation Learning For \textit{P}erson \textit{R}e-identification (termed as MDPR), which addresses the challenging problem from multiple perspectives within a single unified model, leveraging the power of mutual distillation to enhance the feature representations collectively. Specifically, our approach encompasses two branches: a hard content branch to extract local features via a uniform horizontal partitioning strategy and a Soft Content Branch to dynamically distinguish between foreground and background and facilitate the extraction of multi-granularity features via a carefully designed attention mechanism. To facilitate knowledge exchange between these two branches, a mutual distillation and fusion process is employed, promoting the capability of the outputs of each branch. Extensive experiments are conducted on widely used person ReID datasets to validate the effectiveness and superiority of our approach. Notably, our method achieves an impressive $88.7\%/94.4\%$ in mAP/Rank-1 on the DukeMTMC-reID dataset, surpassing the current state-of-the-art results. Our source code is available at https://github.com/KuilongCui/MDPR.
\end{abstract}


\section{Introduction}

\IEEEPARstart{P}{erson} re-identification (ReID) is aimed at re-identifying the person of interest under different cameras. It has attracted a significant amount of attention in computer vision and machine learning communities due to its broad applications in public security, surveillance, and social media analysis. Person re-identification is a challenging problem due to variations in pose, illumination, occlusion, and camera viewpoint \cite{zheng2015scalable, li2014deepreid, wei2018person}.

\begin{figure}[thb]
\centering
\includegraphics[width=\linewidth]{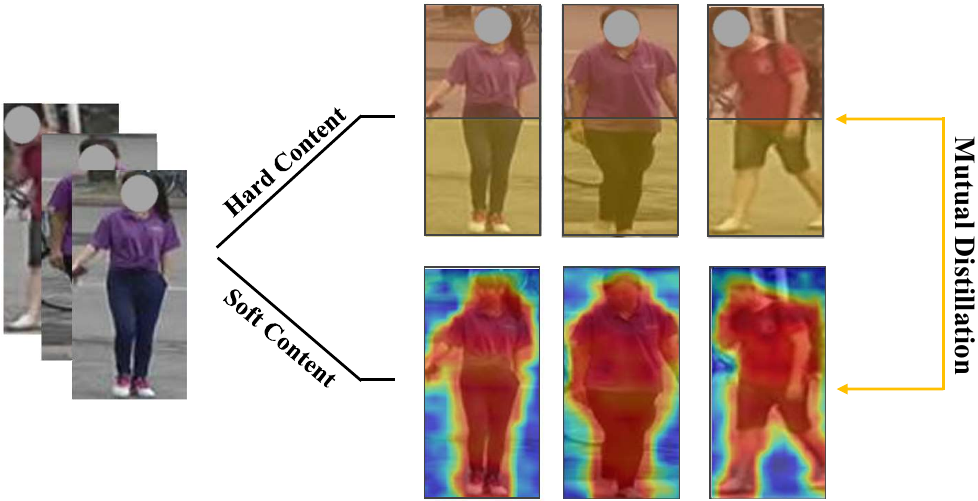}
\caption{The overview structure of our method. The upper part corresponds to the Hard Content Branch, which uniformly partitions all images horizontally into two parts. The lower part represents the Soft Content Branch, which leverages attention mechanisms to distinguish individuals from the background. The two branches engage in mutual distillation to enhance their respective feature representation capabilities.}
\label{introduce method}
\end{figure}

Therefore, in recent years, a number of methods have been proposed to address this challenging problem, and the field of person re-identification research has witnessed significant advancements \cite{zheng2017unlabeled, wei2018person, luo2019bag, luo2021self}. In the existing literature, most existing person re-identification methods are primarily designed to focus on extracting discriminative features, which can be divided into two groups, attention-based methods \cite{chen2019abd,chen2021person,zhang2020relation,chen2019self} and stride-based methods \cite{sun2018beyond,wang2018learning,fu2019horizontal,zheng2019pyramidal,zhang2021person}.

Attention mechanisms, which aim to highlight salient regions in an input, demonstrate considerable promise for person re-identification. By dynamically modeling the importance of different spatial areas within an image, attention can facilitate the focus on discriminative body parts for person matching across camera views.This effectively filters out background interference and captures richer detail features. 
Additionally, various techniques, such as attention orthogonality \cite{chen2019abd} and reinforcement learning \cite{chen2019self}, have been introduced to boost the effectiveness of learned attention maps. However, attention mechanisms predominantly concentrate on specific parts with fixed semantics, and variations in pose or occlusion can influence the reliability of representation \cite{wang2018learning}. Moreover, attention generated based on coarse-grained features may be less accurate due to the increased noise inherent in such features.

Another straightforward yet effective method for extracting features is stride partition \cite{sun2018beyond, wang2018learning, fu2019horizontal, zheng2019pyramidal}. In these approaches, images are divided into stripes, and each stripe is treated as a fine-grained local feature. Compared to methods focusing on identifying regions with specific pre-defined semantics for learning local representations, partitioning proves to be efficient and robust in scenarios with significant variances. Additionally, adjusting the number of parts enables the acquisition of discriminative local information across various granularities. However, it's important to acknowledge that this strategy may generate ineffective local features for images that are misaligned or occluded.

We can see that most existing person re-identification methods are primarily designed to extract discriminative features from a single perspective. 
Even in cases where multi-branch networks are employed \cite{wang2018learning, zhu2020aware, zhang2021fpb}, each branch typically follows a similar underlying principle. For example, MGN \cite{wang2018learning} just introduces variations in the number of image stripes as the main difference among branches. 
However, person re-identification often involves complex and diverse scenarios. Approaches that rely solely on a single perspective may excel in some situations but struggle in others. There are some methods that adopt both hard partition and soft attention in their models \cite{li2018harmonious}, but they neglect knowledge transfer between the two strategies, which has the potential to fully leverage the specialized characteristics of both branches and promote comprehensive learning.

In this paper, we introduce a ReID framework that employs both a hard stride-based branch and a soft attention branch in one unified model. The overview structure of our method is shown in Figure. \ref{introduce method}. The first branch, referred to as the Hard Content Branch, employs uniform horizontal partitioning of input images to extract local features. In contrast, the second branch, known as the Soft Content Branch, leverages attention mechanisms to distinguish individuals from the background and identify discriminative regions within the human body. 
Moreover, we introduce a knowledge distillation and fusion module that facilitates mutual distillation between the branches and aggregates the results from each branch.

The Hard Content Branch employs a uniform partitioning approach that is independent of image content, hence its designation as \textit{Hard}. In contrast, the Soft Content Branch dynamically determines the location of the human body within images, justifying its name as \textit{Soft}. While the Hard Content Branch predominantly focuses on fine-grained local details, the Soft Content Branch is inclined towards capturing global features across multiple granularities. Despite the different starting points for feature extraction in these two branches, the features derived from both branches should exhibit consistency because they both extract features from the same individual. Mutual distillation learning helps the Hard Content Branch learn fine-grained information about the target persons while enabling the Soft Content Branch to focus on local details and the fusion module integrates features extracted by both methods to further extract effective features. Additionally, our model is designed to be end-to-end, simplifying both the training and deployment processes.

It's important to note that this paper serves as an extension of our prior conference paper \cite{an2022attention},  which employs multiple branches to extract local features and attention-enhanced features. In this extended version, we introduce mutual distillation learning between branches to strengthen feature representations for each branch. And the Attention Branch now incorporates attention guidance and enables the extraction of multi-granularity features. Moreover, the features from each branch are integrated to extract additional valuable information. These modifications substantially improve the model's capacity for feature representation. During the testing phase, we conduct experiments on widely used person re-identification benchmark datasets. This allows us to provide more comprehensive and qualitative results along with an extensive ablative analysis, which demonstrate the effectiveness of each component within our proposed framework.

In conclusion, our contributions can be summarized as follows:
\begin{enumerate}
    \item We present a dual-branch framework that exploits heterogeneous cues from two perspectives for feature extraction within a single model. The Hard Content Branch employs a uniform horizontal partition to extract fine-grained local features, and meanwhile, the Soft Content Branch utilizes attention mechanisms to extract global features at multiple granularities.
    \item A mutual distillation and fusion process enables knowledge exchange between the two branches, where the Hard Content Branch imparts local details to the Soft Content Branch, while the latter conveys fine-grained details and positional information back to the former. 
    \item Comprehensive experiments are conducted on mainstream person ReID datasets to demonstrate the superiority of our approach, surpassing the state-of-the-art performance.
\end{enumerate}

\section{Related Works}
In recent years, person re-identification technology has witnessed rapid advancements, with numerous methods being proposed to image and video related field \cite{9351755,9052709,8954105,8621027,qi2019sports,lv2023disentangled}. Following the strong baseline \cite{luo2019strong}, we will introduce various approaches, including Global-based, Attention-based, Pose-based, Mask-based, and Stride-based methods. Additionally, we will provide a brief overview of distillation learning.

\subsection{Global-based Approach}
Global-based approaches typically employ efficient backbone networks to learn identity features, with a focus on data augmentation, loss function design, and training techniques for improving performance. In general, these methods\cite{luo2019bag, he2020fastreid} are capable of achieving efficient but limited results.

Data augmentation techniques are commonly used to diversify training data and enhance model performance. For example, Random Erasing (REA) \cite{zhong2020random} randomly erases a rectangular region in an image to encourage the network to learn more robust features. Random Patch (RPT) \cite{zhou2019omni} is another technique that uses a region from other images in the dataset to cover a region of the target image, improving generalization to occlusions and other forms of variability.

For person re-identification, two types of loss functions are commonly used: classification loss function and triplet loss function. Classification loss treats person ReID as a classification problem where each person is treated as a separate ID\cite{zheng2017person}. Triplet loss treats it as a clustering problem that aims to increase the distance between sample point and negative samples\cite{hermans2017defense}.

Training tricks also make a significant contribution to improving the final performance. For example, BNNeck\cite{he2020fastreid} is introduced to resolve the conflict between the ID loss function and the Triplet loss function during training. Warming up learning rate\cite{fan2019spherereid}, label smoothing\cite{szegedy2016rethinking}, and last stride=1\cite{sun2018beyond} are additional techniques utilized to enhance the performance by more than $10\%$\cite{he2020fastreid} in mean average precision (mAP).

\subsection{Attention-based Approach}
The attention mechanism has shown great effectiveness in computer vision and is widely used in person re-identification. ABD-Net\cite{chen2019abd} introduces a pair of complementary attention modules that focus on channel aggregation and position awareness, respectively. And an orthogonality constraint is also introduced to promote diversity in both hidden activations and weights. In AP-Net\cite{chen2021person}, features are split into multiple local parts. And the corresponding local attention is merged and stacked with the residual connection as an attention pyramid to exploit attention regions in a multi-scale manner, mimicking human visual perception. RGA-SC\cite{zhang2020relation} proposes to stack pairwise correlations of each feature position and the feature itself to learn attention and obtain a compact understanding of global structural information and local appearance information. SCAL\cite{chen2019self} attempts to learn attention by estimating the quality of attention maps in a reinforcement learning manner during the learning process, trying to make it more effective than weakly supervised training approaches employed by many existing methods.

Compared to many of the attention mechanisms currently employed, our approach to attention generation in the Soft Content Branch is relatively straightforward, primarily involving several convolutional operations. Furthermore, we introduce an attention guidance mechanism to further enhance the quality of the attention maps generated for coarse-grained features.

\subsection{Pose-based, and Mask-based Approach}
Recently, pose estimation\cite{xu2022vitpose, cai2020learning} and human semantic parsing\cite{li2017multiple} are largely improved in accuracy. Typically, these models identify skeletal points of individuals\cite{li2019pose} or generate masks of body parts\cite{guo2019beyond} to extract essential location information. It is reasonable to use pose estimation and human semantic parsing for handling pose variations and background clutters in ReID, and such attempts have obtained a great improvement in performance\cite{li2019pose}. 

AACN\cite{xu2018attention} deeply exploits the pose information and uses fourteen key points of the human body to guide the learning of part attention, which mask out undesirable background features in feature maps to deal with part occlusion. SPReid\cite{kalayeh2018human} generates probability maps through the human semantic parsing module. DSA\cite{zhang2019densely} uses the DensePose\cite{guler2018densepose} to estimate the dense semantics of a 2D image and constructs a set of densely semantically aligned part images, which act as a regulator to guide the model to learn densely semantically aligned features from the original image. As there exists many useful contextual cues that do not fall into the scope of predefined human parts or attributes, $p^{2}$-Net\cite{guo2019beyond} apply a human parsing model\cite{ruan2019devil} to extract the binary human part masks and a self-attention mechanism to capture the soft latent (non-human) part masks. ISP\cite{zhu2020identity} is another approach that locates both human body parts and personal belongings at a pixel level for aligned person ReID only with person identity labels.

The present state of off-the-shelf pose estimation and human parsing methods indicates that they are primarily tailored to specific scenarios, leading to a lack of generalizability in person ReID. This can be attributed to significant disparities in the datasets utilized for training, including variations in resolution, image quality, illumination, and other factors. Moreover, the absence of semantics or pose labels for ReID datasets complicates the process of generating precise critical points or masks for the human body in the context of person ReID.

\subsection{Stride-based Approach}
Stride-based methods partition images into several parts and learn local features from each part, which are then aggregated to obtain the final feature representation.

PCB\cite{sun2018beyond} is the first stride-based deep learning method for person ReID. In addition to partitioning images, they also use an addition module to reassign outliers to the closest part to enhance consistency within each part. MGN\cite{wang2018learning} proposes a multi-branch deep network architecture consisting of one branch for global feature representation and two branches for local feature representation, which divides the image into different numbers of strips to obtain a local feature representation. HPM\cite{fu2019horizontal} uses partial feature representations at different horizontal pyramidal scales for classification, enhancing the discrimination of various character parts. Pyramid\cite{zheng2019pyramidal} proposes a coarse-to-fine pyramid model that not only incorporates local and global information but also integrates the gradual cues between them. HLGAT\cite{zhang2021person} uses the local features obtained from horizontal partitioning to construct a graph for modeling inter-local and intra-local relationships, achieving impressive performance. Despite the efforts of stride-based approaches, they inevitably fall into the dilemma of unaligned images and occlusion.

\subsection{Distillation learning}
Model distillation is an effective and widely used technique to transfer knowledge from a teacher to a student network\cite{ba2014deep}. In terms of representation of the knowledge to be distilled, existing models typically use teacher’s class probabilities \cite{hinton2015distilling} and/or feature representation \cite{romero2014fitnets}. And, some methods have proposed an alternative approach where an ensemble of students teaches each other through mutual distillation \cite{zhang2018deep}.

For current distillation techniques, the emphasis is placed on utilizing multiple backbones to facilitate knowledge transfer. In contrast, our approach employs a single backbone network with two distinct branches. Significantly, these two branches extract features from different perspectives, ensuring the effectiveness of knowledge transfer.

\section{Proposed Method}

\textbf{Problem Formulation:}~Given a dataset of $n$ images $I = \{I_i\}_{i=1}^{n}$ representing individuals from $n_{id}$ distinct people captured by non-overlapping camera views, with corresponding identity labels $Y = \{y_i\}_{i=1}^{n} ( y_i \in [1,\ldots, n_{id}])$,  the objective is to learn a deep feature representation model for person re-identification matching. To this end, we propose a dual-perspective feature extraction model enhanced through mutual distillation methodology. 

\begin{figure*}[ht]
\centering
\includegraphics[width=\linewidth]{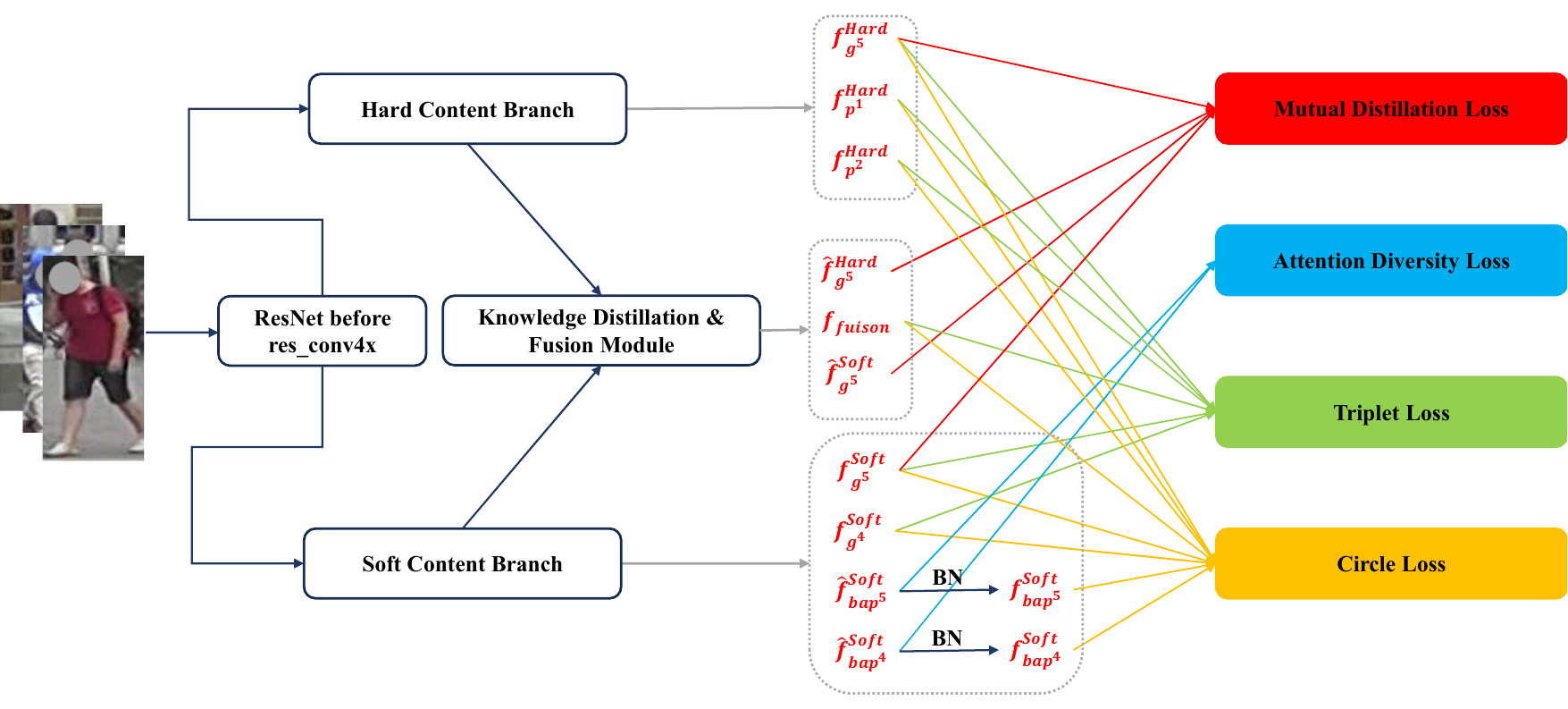}
\caption{The overall architecture of our proposed network, which consists of two branches and a knowledge distillation and fusion module. The Hard Content Branch applies a uniform partition to the input image, while the Soft Content Branch distinguishes between individuals and the background based on attention. The knowledge distillation and fusion module facilitates mutual distillation learning and integrates the outputs of both branches to further extract meaningful information. BN denotes the batch normalization operation.}
\label{network}
\end{figure*}

\textbf{Framewok Overview:}~As shown in Figure.~\ref{network}, the overall structure of our proposed network includes two branches and a knowledge distillation and fusion module. We adopt ResNet as the backbone network, with two branches sharing the low-level network while independently utilizing the high-level network. These branches are tailored to extract different features using varied techniques, while the knowledge distillation and fusion module conducts mutual distillation and integrates the outputs of both branches to extract meaningful information. The algorithm utilized in the training process is explicated in Algorithm. \ref{Traing Procedure}, adopting an end-to-end framework that offers a convenient approach to training and $f^{Hard}_{All}$, $f^{Soft}_{All}$, and  $f_{fusion}$ correspond to all features generated by the Hard Content Branch, all features produced by the Soft Content Branch, and the fused output from the Knowledge Distillation and Fusion Module. Each component in our model plays a crucial role in the overall architecture and we will describe how these components perform in the next subsections.

\begin{algorithm}
    \label{Traing Procedure}
    \caption{Training Our Proposed Network}
    \KwIn{Training dataset $\{(I_i, y_i)\}_{i=1}^{n}$, Pretrained weights ${\theta}^{\prime}$, Learning rate $\eta$, Number of epochs $E$, Target network $\mathcal{M}(\theta)$.}
    \KwOut{Trained network $\mathcal{M}$.}
        Initialize $\mathcal{M}$ to ${\theta}^{\prime}$;
        
        \For{$epoch \leftarrow 1$ \KwTo $E$}{
            \For{each mini-batch $(x, y)$ in training data}{
                $f^{Hard}_{All}, f^{Soft}_{All}, f_{fusion} = \mathcal{M}(x)$;
                
                Calculate the loss $\mathcal{L}$ over the object function \ref{loss function total}\;
                
                Compute the gradient of the loss $\nabla_{\theta} \mathcal{L}$;
                
                Update model parameters: $\theta \leftarrow \theta - \eta \nabla_{\theta} \mathcal{L}$\;
            }
        }
        
        \KwRet{$\mathcal{M}$}
\end{algorithm}

\subsection{Hard Content Branch}
The Hard Content Branch utilizes a uniform horizontal partition to extract multiple discriminative local features. For each input image, the partitioning process remains consistent, not accounting for the uniqueness of the image content.

The output of the \emph{res-conv5} in this branch is defined as $C^{Hard}_5 \in \mathbb{R}^{C \times H \times W}$ where \emph{C}, \emph{H} and \emph{W} represent the number of channels, height and width, respectively. Then, we uniformly divide $C^{Hard}_5$ into \emph{N} segments along the height dimension. Many prior studies \cite{sun2018beyond, wang2018learning, fu2019horizontal, zhang2021person, an2022attention} have employed similar approaches. In this study, considering that the majority of images in person re-identification datasets \cite{zheng2015scalable,ristani2016performance} are aligned, we define \emph{N} = 2 by experience to extract critical features of the target pedestrians based on the upper and lower halves of the human body. The partitioned features are denoted as $\{P_{1}^{Hard},P_{2}^{Hard}\}$, where $P_{i}^{Hard} \in \mathbb{R}^{C \times (H/2) \times W}$  contains localized information from a specific region.

\begin{figure}[ht]
\centering
\includegraphics[width=\linewidth]{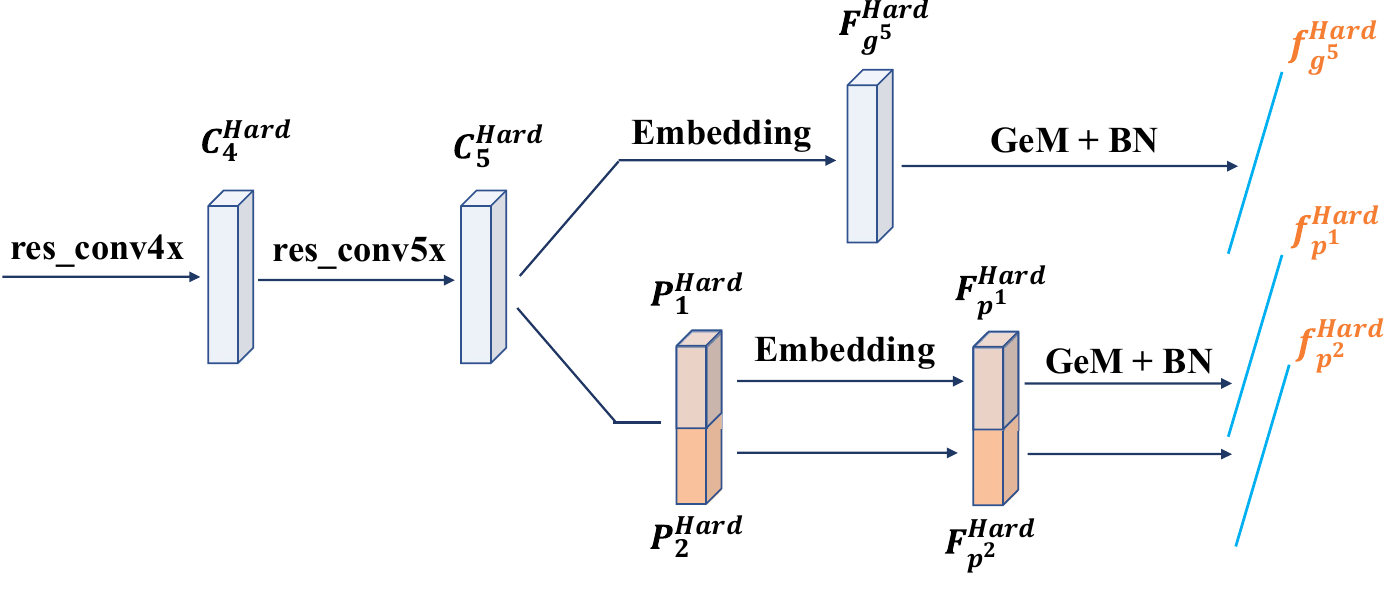}
\caption{The architecture of Hard Content Branch. In our experiments, we partition the features into two segments. The Embedding block is composed of a 1x1 convolution, a batch normalization layer, and a relu activation function. GeM refers to Generalized Mean Pooling. BN refers to batch normalization operation.}
\label{hard_content_branch}
\end{figure}

In order to enhance the extraction of crucial contextual information and reduce the feature dimensionality, an embedding block, which typically includes a convolutional layer, a batch normalization (BN) layer, and a relu activation function, is employed to the backbone output and each partitioned feature. After embedding, we perform Generalized Mean Pooling (GeM) for each feature. Define $F \in \mathbb{R}^{M \times H \times W}$ as the feature after embedding where $M$ is the embedding dimension, $H \times W$ is the feature map size, and $f \in \mathbb{R}^{M}$ as the output of GeM. The GeM operation can be formulated as follows:
\begin{equation}
f_{i}=\left(\frac{1}{\left|{F}_{i}\right|} \sum_{f \in {F}_{i}} f^{\alpha}\right)^{\frac{1}{\alpha}},
\end{equation}
where $f_{i}$ is the $i$-th feature point in $f$ and $F_{i}$ refers to the $i$-th feature maps in $F$. The pow operation parameter $\alpha$ is trainable and is initialized to 3. At last, a batch normalization layer is applied to each feature maps to produce the final output. Specifically, for each partitioned feature, the Embedding, GeM, and BN do not share parameters. Overall, the Hard Content Branch generates one Global feature $f^{Hard}_{g^5}$and two Partition features $\{f^{Hard}_{p^1},f^{Hard}_{p^2}\}$. 

\subsection{Soft Content Branch}
In contrast to the Hard Content Branch, Soft Content Branch generates unique attention maps based on the content of each image to guide the extraction of discriminative features. To further enhance the model's feature representation capabilities, we employ multi-granularity attention to boost features at corresponding granular levels. In addition, attention maps derived from fine-grained features are employed to steer the generation of attention maps for coarse-grained features, mitigating the impact of noise in the latter.

Before presenting the specific design, we first describe how to generate the attention maps. Given an input feature ${X} \in \mathbb{R}^{C \times H \times W}$, where C, H, and W represent the number of channels, height, and width respectively, we employ the module $\phi$ to generate raw attention:

\begin{equation}
    A = \phi ( X ; W_{\phi} ) \in \mathbb{R}^{ K \times H \times W},
\end{equation}
where \emph{K} is the number of generated attention maps and $W_{\phi}$ represents the parameters of model $\phi$.

\begin{figure}[ht]
\centering
\includegraphics[width=\linewidth]{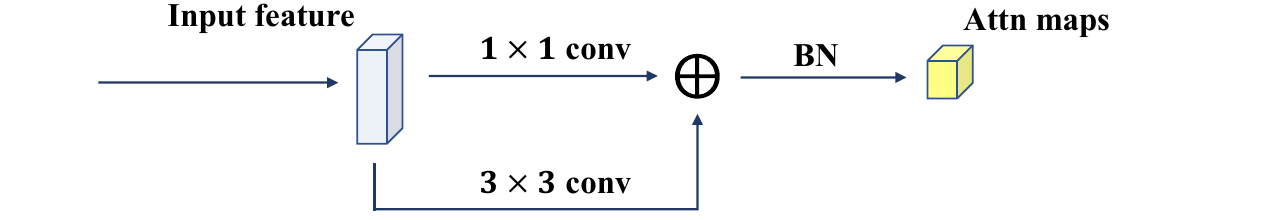}
\caption{The structure of attention generation module $\phi$. $\oplus$ denotes the element-wise sum.}
\label{attn generation}
\end{figure}

The module $\phi$ consists of a 1x1 convolution, a 3x3 convolution, a batch normalization layer, and a relu activation function. As illustrated in Figure. \ref{attn generation}, the outputs of the two convolutions undergo element-wise summation, followed by batch normalization and relu operations. Consequently, \emph{K} attention maps are produced, each responsive to distinct human body parts.

To effectively distinguish between individuals and the background and increase the variance of attention maps, following the method proposed by \cite{wang2023learning}, we generate \emph{K+1} attention maps and apply the softmax operation along the channel dimension of the attention maps, subsequently choosing the top \emph{K} maps as the ultimate output. This leads to the generation of the last attention map that represents the background, while the remaining ones focus on body parts.

\begin{figure}[ht]
\centering
\includegraphics[width=\linewidth]{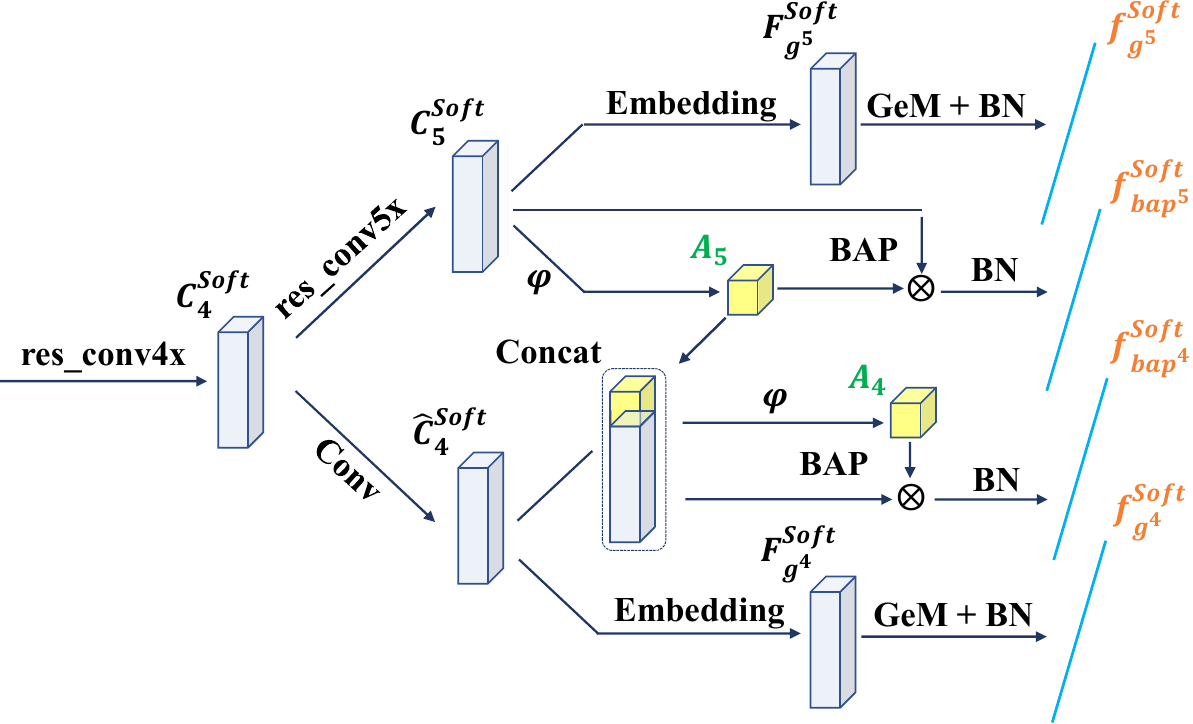}
\caption{The architecture of Soft Content Branch. $\phi$ refers to the attention generation module. BAP refers to the Bilinear Attention Pooling. Conv refers to a 1x1 convolution layer.}
\label{soft_content_branch}
\end{figure}

The overall architure of the Soft Content Branch is shown in Figure. \ref{soft_content_branch}. In this branch, we utilize the feature activations output by the last two stage's residual block, denoted as $C^{Soft}_4$, $C^{Soft}_5$ for \emph{res-conv4} and \emph{res-conv5} outputs. 
Initially, the generation module $\phi$ is employed to create attention maps $A^5$ based on $C^{Soft}_5$. Subsequently, Bilinear Attention Pooling (BAP) \cite{hu2019see} is applied, followed by a batch normalization operation on the BAP result. The process can be expressed as:
\begin{equation}
\hat{{f}}^{Soft}_{{bap}^{5}}
= 
    \left(
        \begin{array}{c}
            {\hat{f}^{soft}_{{bap}^5_1}} \\ 
            {\hat{f}^{soft}_{{bap}^5_2}} \\ 
            \vdots \\
            {\hat{f}^{soft}_{{bap}^5_K}}
        \end{array}
    \right)
= 
    \left(
        \begin{array}{c}
            GeM(EO({{A}_{1}^{5}} \odot {C^{Soft}_5}))) \\
            GeM(EO({{A}_{2}^{5}} \odot {C^{Soft}_5}))) \\
            \vdots \\
            GeM(EO({{A}_{K}^{5}} \odot {C^{Soft}_5})))
        \end{array}
    \right),
\label{BAP}
\end{equation}

\begin{equation}
{f}^{Soft}_{{bap}^{5}} = BN(\hat{{f}}^{Soft}_{{bap}^{5}}),
\end{equation}
where $\odot$ denotes the element-wise multiplication, ${{A}_{i}^{5}}$ refers to the \emph{i}-th attention map of  ${{A}^{5}}$, and \emph{EO} denotes the embedding operation. Specifically, the embedding and GeM operations are not parameter-shared across the different attention maps.

Then, we employ ${A^5}$ to guide the generation of ${A^4}$. As shown in Figure. \ref{soft_content_branch}, we subject ${C^{Soft}_4}$ to a convolution operation, followed by its concatenation with ${A^5}$ and the attention maps ${A^4}$ corresponding to ${C^{Soft}_4}$ are generated based on this concatenation. Then, the BAP and batch normalization operation is performed to generate feature maps $\hat{f}^{Soft}_{{bap}^{4}}$ and ${f}^{Soft}_{{bap}^{4}}$. In addition, the embedding, GeM, and batch normalization operations are applied to $\hat{C}^{Soft}_4$ and ${C^{Soft}_5}$, resulting in the generation of $f^{Soft}_{g^4}$ and $f^{Soft}_{g^5}$.

\subsection{Knowledge Distillation and Fusion Module}
The Knowledge Distillation and Fusion Module conducts mutual distillation learning based on the output features from the Hard Content Branch and the Soft Content Branch and integrates the features from both branches to extract additional valuable information.

We perform mutual distillation based on $f^{Hard}_{g^5}$ from the Hard Content Branch and $f^{Soft}_{g^5}$ from the Soft Content Branch. A Multi-Layer Perceptron layer (MLP), which comprises two linear layers, a batch normalization layer, and a relu activation function, is responsible for transforming the features from one domain to match the characteristics of the other domain. Denote $\hat{f}^{Hard}_{g^5}$ and $\hat{f}^{Soft}_{g^5}$ as the results of $f^{Hard}_{g^5}$ and $f^{Soft}_{g^5}$ when fed into the MLP and we conduct distillation learning based on the cosine similarity between $\hat{f}^{Soft}_{g^5}$ and $f^{Hard}_{g^5}$, as well as $\hat{f}^{Hard}_{g^5}$ and $f^{Soft}_{g^5}$. Detailed information about the loss design is provided in Section \ref{loss} Mutual Distillation Loss.

To fuse multiple features, a fusion module is introduced, comprising a 1x1 convolution, a 3x3 convolution, a batch normalization layer, and a relu activation function. Initially, we merge the features within each branch, and then further integrate the fusion results obtained from each branch to obtain the final fused features. Subsequently, we execute GeM and BN to obtain the final fused feature $f_{fusion}$.

\begin{figure}[ht]
\centering
\includegraphics[width=\linewidth]{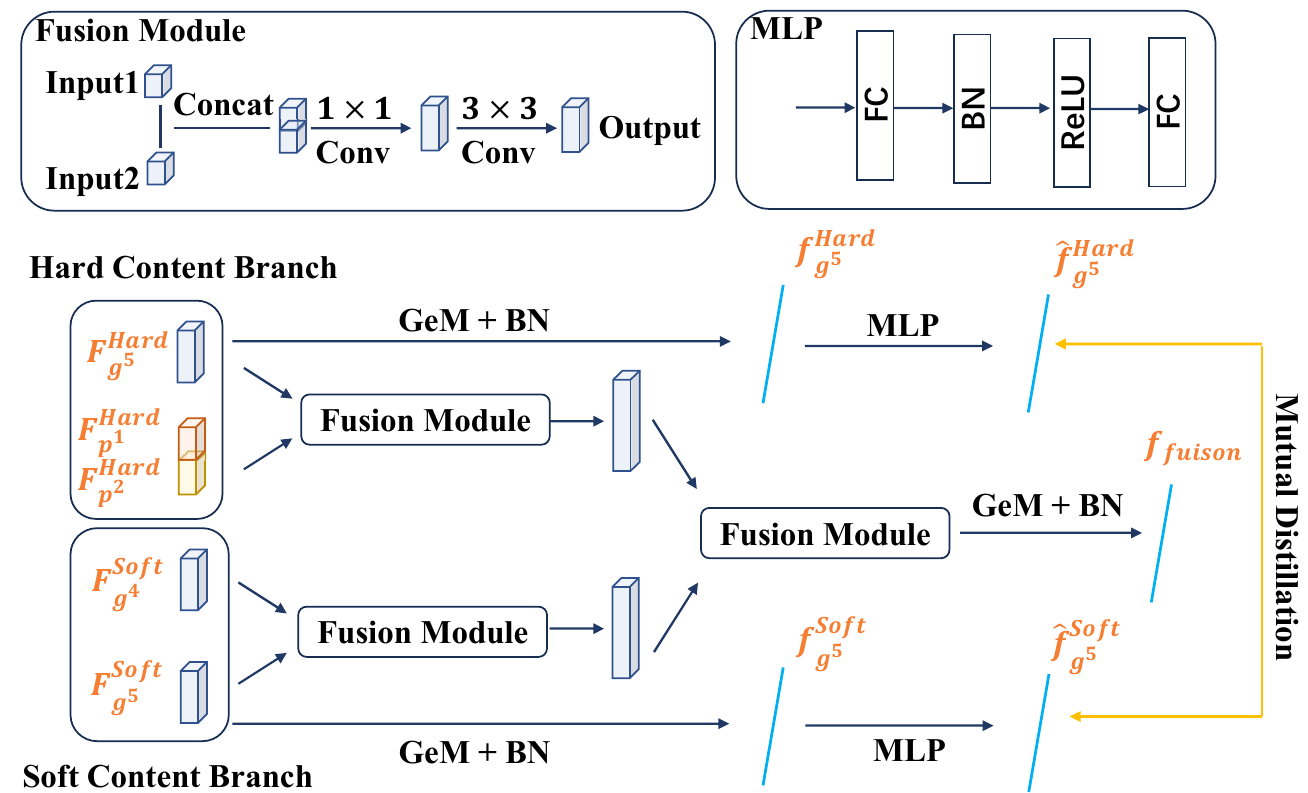}
\caption{The architecture of Knowledge Distillation and Fusion Module. In particular, the inputs consist of features output by the embedding blocks of both the Hard Content Branch and the Soft Content Branch.}
\label{distillation_and_fusion}
\end{figure}

\subsection{Objective Function}\label{loss}
To enhance the feature representation within our model, we integrate both the circle loss and triplet loss as our loss functions. Moreover, to augment the diversity and effectiveness of the attention maps generated in the Soft Content Branch, we introduce a specialized loss dedicated to fostering attention exclusivity. Then, a mutual distillation learning loss is employed to facilitate knowledge transfer between the Hard Content Branch and the Soft Content Branch, further fortifying the model's feature representation capabilities. 

\textbf{Circle loss}\cite{sun2020circle}. Circle loss assigns a distinct penalty strength to each similarity score based on its distance to the optimal effect, which allows it to optimize each similarity score with a different gradient, leading to better overall performance. We calculate the similarity scores between anchor feature $x$ and the other features in a mini-batch. Specifically, we denote $s_{n}^{j}=x_{j}^{T} \cdot x /(\|x_{j}\|_{2} \cdot \|x\|_{2})$ where $x_{j}$ is the $j$-th sample in the negative sample set $\mathcal{N}$ and $s_{p}^{i}=x_{i}^{T} \cdot x /(\|x_{i}\|_{2} \cdot \|x\|_{2})$ where $x_{i}$ is the $i$-th sample in the positive sample set \(\mathcal{P}\). Correspondingly, $T=|\mathcal{P}|, L=|\mathcal{N}|$. The circle loss can be formulated as:

\begin{equation}
\begin{split}
    \mathcal{L}_{\text{circle}}=\log
    [1+ 
    \sum_{j=1}^{L} \exp \left(\gamma \alpha_{n}^{j}\left(s_{n}^{j}-\Delta_{n}\right)\right) \\ \sum_{i=1}^{T}\exp \left(-\gamma \alpha_{p}^{i}\left(s_{p}^{i} -\Delta_{p}\right)\right)
    ],
\end{split}
\end{equation}
where $\alpha_{p}^{i}=\left[O_{p}-s_{p}^{i}\right]_{+}$, $\alpha_{n}^{j}=\left[s_{n}^{j}-O_{n}\right]_{+}$, and $[\cdot]_{+}$is the "cut-off at zero" operation to ensure $\alpha_{p}^{i}$, $\alpha_{n}^{j}$ are non-negative. $O_{p}=1+m, O_{n}=-m, \Delta_{p}=1-m$, and $\Delta_{n}=m$. There are two hyper-parameters: the scale factor $\gamma$ and the relaxation margin $m$. Following SBS \cite{he2020fastreid}, we set $\gamma$ to 64 and $m$ to 0.35 in experiments.

\textbf{Triplet loss} \cite{hermans2017defense}. As a classic metric learning loss, triplet loss enhances the robustness and further improve the performances at the same time. The triplet loss with a margin hyper-parameter $\rho$ can be formulated as:
\begin{equation}
\begin{split}
   \mathcal{L}_{\text{triplet}}=-\sum_{i=1}^{R} \sum_{a=1}^{S}
    [\rho + \max _{p=1 \ldots R}\left\|{f}_{a}^{(i)}-{f}_{p}^{(i)}\right\|_{2} 
    \\ -\min _{\substack{n=1 \ldots  S\\ j=1 \ldots R \\ {j \neq i}}}
    \left\|{f}_{a}^{(i)}-{f}_{n}^{(j)}\right\|_{2}],
\end{split}
\end{equation}
where ${f}_{a}^{(i)}$, ${f}_{p}^{(i)}$ and ${f}_{n}^{(j)}$ denote the anchor feature, the positive feature and  the negative feature which come from the \emph{i}-th identity for a mini-batch with $R$ selected identities and $S$ images from each identity. The triplet loss function encourages the model to learn feature representations such that the distance between the anchor and positive samples is smaller than the distance between the anchor and negative samples by at least the margin parameter $\rho$. We set $\rho$ to 0.05 in experiments.

\textbf{Attention Diversity loss}. We employ a straightforward convolutional approach to generate the required attention for the BAP operation. The utilization of softmax function effectively enhances the model's ability to distinguish between foreground and background. However, the attention maps focused on the foreground tend to concentrate on the same key regions, resulting in a substantial similarity among the generated attention maps. 

To encourage the generated attention maps to focus on distinct key regions, given the feature $f = ({f_1},\dots,{f_i},\dots, {f_K})$ where $f_i$ is the pooled result from element-wise multiplying the $i$-th attention map with the feature along the channel dimension, and \emph{K} refers to the number of attention maps. We define a similarity metric between the results of Bilinear Attention Pooling (BAP). Additionally, a regularization term is introduced to mitigate the risk of overfitting. The attention diversity loss can be formulated as:

\begin{equation}
    \mathcal{L}_{\text{attention}} = \sum_{\substack{i=1, \ldots  ,K\\ j=1, \ldots ,K \\ {j \neq i}}}
    \frac{{f_i} \cdot {f_j}}{||{f_i}||_{2} \cdot ||{f_j}||_{2}} + \beta \cdot \sum_{i=1}^{K} ||{f_i}||_{1},
\end{equation}
where $\beta$ is a hyper-parameter, and is to 0.001 in our experiments. 

\textbf{Mutual Distillation loss}. The Hard Content Branch and the Soft Content Branch extract features from different domains, and we utilize features generated by both branches for mutual distillation learning, aiming to achieve the transfer of knowledge. Denote $\mathcal{D}(p, z) = - ({p \cdot z}) / ({|| p ||_2 \cdot || z ||_2 })$ as the negative cosine similarity between vectors $p$ and $z$, and the mutual distillation loss can be formulated as:
\begin{equation}
\begin{split}
    \mathcal{L}_{\text{distillation}} = \frac{1}{2}( \mathcal{D}(\hat{f}^{Hard}_{g^5}, stopgrad({f}^{Soft}_{g^5})) + \\ \mathcal{D}(\hat{f}^{Soft}_{g^5}, stopgrad({f}^{Hard}_{g^5})) ),
\end{split}
\end{equation}
where $stopgrad$ means preventing the backpropagation of gradients.

Finally, our model uses a combination of these four loss functions, which can be expressed as:
\begin{equation}
 \mathcal{L} =  \mathcal{L}_{\text{circle}} +  \mathcal{L}_{\text{triplet}} +  \mathcal{L}_{\text{attention}} +  \mathcal{L}_{\text{distillation}}.
 \label{loss function total}
\end{equation}

\subsection{Inference Strategy}

During the inference phase, we combine the outputs from the Hard Content Branch, Soft Content Branch, and the fusion results in Knowledge Distillation and Fusion Module to acquire the final feature matrix. The dimensionality of each feature, along with its relevance for inference, is detailed in Table. \ref{feature dimension}. Specifically, the features obtained from BAP are not included in the references and all the features used in reference are concatenated. We calculate the cosine distances between the query image and all images in the gallery, subsequently ranking them to obtain the retrieval results. 

\begin{table}
\begin{center}
\caption{The dimensionality of generated features and their relevance in inference. $M$ refers to the embedding size and $K$ refers to the number of attention maps used in Soft Content Branch.}
  \begin{tabular}{ccc}
    \toprule
    Feature & Dimsion & Used in Inference\\
    \midrule
    $f^{Hard}_{g^5}$& $M$  & \checkmark \\
    $f^{Hard}_{p^{1,2}}$& $M$  & \checkmark \\
    $f^{Soft}_{g^5}$& $M$  & \checkmark\\
    $f^{Soft}_{g^4}$& $M$  &\checkmark\\
    $\hat{f}^{Soft}_{bap^5},f^{Soft}_{bap^5}$& $M \times K$&\\
    $\hat{f}^{Soft}_{bap^4},f^{Soft}_{bap^4}$& $M \times K$&\\
    $f_{fusion}$& $4 \times M$&\checkmark\\
    \bottomrule
  \end{tabular}
  \label{feature dimension}
\end{center}
\end{table}

\section{Experiment}
\begin{figure}[ht]
\centering
\includegraphics[width=\linewidth]{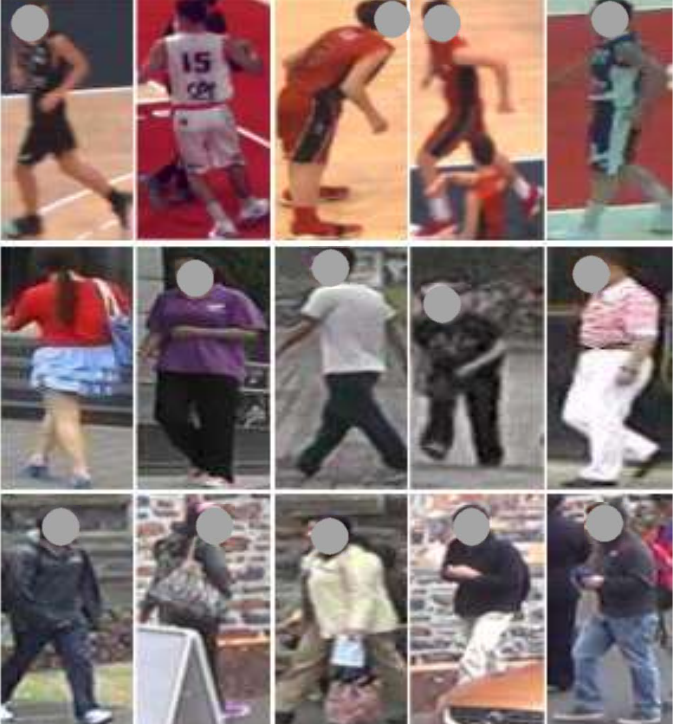}
\caption{Examples of images from the candidate datasets. The first row contains samples from SynergyReID, the second row from Market-1501, and the third row from DukeMTMC-reID.}
\vspace{-3mm}
\label{datasets}
\end{figure}

In order to evaluate the effectiveness of our model and establish comparisons with other approaches, we conduct experiments on one player re-identification dataset and two person re-identification datasets. And some samples in the candidate datasets are shown in Figure. \ref{datasets}.

\subsection{Dataset and Evaluation}

\textbf{SynergyReID} \cite{zandycke2022deepsportradarv}. This dataset is built using 99 short video sequences from 97 different professional basketball games of the LNB proA league played in 29 different arenas. It encompasses a wide variety of players and sportswear appearances, capturing the dynamic nature of the game across multiple arenas with varying illumination and court layouts. It includes 9,529 image crops of players, referees and coaches. The training set consists of 8,569 images representing 436 individuals, and the testing set is further categorized into 50 query images and 910 gallery images.

\textbf{Market-1501} \cite{zheng2015scalable}. This dataset is collected in front of a supermarket at Tsinghua University. The dataset contains 32,668 images of 1,501 individuals captured by six cameras and each individual in the dataset is captured by at least two cameras. The whole dataset is divided into a training set with 12,936 images of 751 persons, a testing set with 3,368 query images and 19,732 gallery images of 750 persons.

\textbf{DukeMTMC-reID} \cite{ristani2016performance}. This dataset contains a total of 36,411 images of 1,812 pedestrians captured by eight high-resolution cameras in an outdoor environment. Among the 1,812 pedestrians, 1,404 of them are captured by more than two cameras, while the remaining 408 are captured by only one camera. The dataset is partitioned into three distinct sets: the training set, the test set, and the query set. The training set encompasses 702 pedestrians, comprising a total of 16,522 images. In the test set, there are 702 pedestrians, alongside an additional 408 interfering pedestrians, yielding a combined count of 17,661 images. The query set encompasses 702 pedestrians, randomly extracted from the test set, with the selection of one image from each camera for every pedestrian. This results in a total of 2,228 images in the query set.

\textbf{Evaluation Metrics.} To evaluate our model's performance and compare it with other methods, we report the mean average precision (mAP) and cumulative matching characteristics (CMC) at Rank-1 on all the candidate datasets.

\subsection{Implementation}

Our model incorporates a modified ResNet-101 as its backbone due to the excellent performance it has demonstrated in ReID models\cite{luo2019bag,sun2018beyond}. Specifically, in our model the backbone parts before \emph{res-conv4} are shared by each branch, while \emph{res-conv4} and \emph{res-conv5} are not shared by each branch. In order to preserve more fine-grained details in the feature maps, we modify the stride for the last down-sampling operation in both branches from 2 to 1, reducing the amount of spatial information lost during the down-sampling operation. In addition, we employ IBN-a\cite{pan2018two} to enhance the modeling ability of the convolutional neural network. While these modifications marginally increase computational cost, the performance benefits are significant\cite{luo2019bag, an2022attention}. 

Considering the inconsistent image size, all the images are uniformly resized to $384 \times 192$ during both the training and testing phases. To make full use of the datasets, data augmentation is employed. Horizontal flipping of images is adopted with a probability of 0.5. Random Erasing (REA)\cite{zhong2020random} and Random Patch (RPT)\cite{zhou2019omni}  are used with a probability of 0.5, which helps to prevent overfitting and increase the robustness of the model. Prior to feeding them to the network, images are normalized by subtracting the mean values [0.485, 0.456, 0.406] and dividing by the standard deviations [0.229, 0.224, 0.225], which helps to improve the convergence speed and accuracy of the training process.

During the training phase, a ResNet-101 model pre-trained on \textit{LUPerson} \cite{fu2021unsupervised} is used as the backbone. The attention number $K$ is set to 2 and the embedding size $M$ is set to 512. To satisfy the requirements of the triplet loss, each batch is sampled with a random selection of 16 identities, and for each identity 16 images are randomly sampled from the training set. In the event that a specific identity does not possess the necessary sixteen images, a random duplication technique is employed to augment the available images until required images are obtained. The weight decay factor is set to 0.0005, and the Adam optimizer is used with a momentum of 0.9. The model is warmed up with a linearly growing learning rate from $3.5 \times 10^{-5}$ to $3.5 \times 10^{-4}$. Then, the learning rate is decreased by CosineAnnealingLR. Considering the difference in dataset sizes, SynergyReID is trained for 40 epochs with 200 warm-up iterations, while Market-1501 and DukeMTMC-reID undergo 60 epochs and 2000 warm-up iterations, respectively.

Our model is implemented using the PyTorch framework, and for training, we utilize Distributed Data Parallel (DDP). All experiments are conducted on 4 NVIDIA GeForce RTX 3090 GPUs.

\subsection{Comparison with State-of-the-Art Methods}

To demonstrate the effectiveness of our proposed approach, we compare it with the current state-of-the-art methods on the candidate datasets. The results show that our proposed method outperforms other state-of-the-art methods or achieves comparable performance on all the candidate datasets.

\begin{table}
\begin{center}
  \caption{Comparisons with the state-of-the-art methods on SynergyReID.}
  \begin{tabular}{cccc}
    \toprule
    Method  &  Publication &  mAP   &   Rank-1  \\ 
    \midrule
    MGN \cite{wang2018learning} & ACM MM18 & 92.5 & 98.0 \\
    BOT \cite{luo2019bag} & CVPRW19 & 89.3 & 98.0 \\
    TransReID \cite{he2021transreid} & ICCV21 & 92.8 & 98.0 \\
    LDS \cite{zang2021learning} & IVC21 & 90.9 & 98.0 \\
    A$^2$MGN \cite{an2022attention}& ACM MMW22& 92.4 & 98.0\\
    SBS \cite{he2020fastreid} & ACM MM23 & 90.3 & 98.0 \\
    MDPR(Ours) & - & \textbf{94.0}& \textbf{98.0}\\ 
    \bottomrule 
  \end{tabular}
  \label{SynergyReID SOTA}
  \vspace{-3mm}
\end{center}
\end{table}

\textbf{SynergyReID.} On the SynergyReID dataset, our method attains $94.0\%$/$98.0\%$ in mAP/Rank-1 accuracy, respectively, outperforming most previously published results. Compared to our prior method, our model employs fewer branches, yet achieves a $1.6\%$ improvement in mAP. Given the relatively small dataset size, it is noteworthy that most methods also achieve a $98.0\%$ Rank-1 accuracy.

\begin{table}
\begin{center}
  \caption{Comparisons with the state-of-the-art methods on Market-1501.}
  \begin{tabular}{cccc}
    \toprule
    Method  & Publication & mAP   &   Rank-1  \\ 
    \midrule
    MGN\cite{wang2018learning} & ACM MM18 & 86.9   & 95.7 \\
    APNet\cite{chen2021person} &  TIP21  & 90.5    &  96.2 \\
    PASS\cite{zhu2022part} & ECCV22 & 93.3    &  96.9 \\
    BPB\cite{somers2023body} & WACV23  & 89.4& 95.7 \\
    MSINet\cite{gu2023msinet} & CVPR23 & 89.6 & 95.3 \\
    SOLIDER\cite{chen2023beyond} & CVPR23 & 93.9 & 96.9 \\
    MDPR(Ours) & - & \textbf{94.2} & \textbf{97.4} \\ 
    \bottomrule
  \end{tabular}
  \label{Market-1501 SOTA}
  \vspace{-3mm}
\end{center}
\end{table}

\textbf{Market-1501.} On the Market-1501 dataset, our approach has yielded satisfactory results. As indicated in Table. \ref{Market-1501 SOTA}, SOLIDER \cite{chen2023beyond} currently stands at the forefront, utilizing prior knowledge from human images to build pseudo semantic labels and import more semantic information into the learned representation. Our approach has achieved mAP and Rank-1 performance metrics of $94.2\%$ and $97.4\%$, respectively. In comparison to SOLIDER \cite{chen2023beyond}, our method outperforms in terms of mAP by $+0.3\%$ and Rank-1 by $+0.5\%$. Furthermore, when compared to networks employing complex attention designs, such as APNet \cite{chen2021person} utilizing the attention pyramid network, our approach demonstrates a significant improvement of $+3.7\%$ in mAP and $+1.2\%$ in Rank-1, showcasing the effectiveness of the attention mechanism we use.

\begin{table}
\begin{center}
  \caption{Comparisons with the state-of-the-art methods on DukeMTMC-reID.}
  \begin{tabular}{cccc}
    \toprule
    Method & Publication &   mAP   &   Rank-1  \\ 
    \midrule
    ISP\cite{zhu2020identity} &   ECCV20&   80.0   &   89.6 \\
    VA-reid\cite{zhu2020aware} &    AAAI20 &    84.5   &   91.6 \\
    HLGAT\cite{zhang2021person} &    CVPR21 &  87.3   &   92.7 \\
    A$^2$MGN\cite{an2022attention} &   ACM MMW22  & 81.4   &   90.0 \\
    GPEOG\cite{li2023effective} & ICME23 & 75.5 & 87.5 \\
    BPB\cite{somers2023body} & WACV23 & 84.2 & 92.4 \\
    MDPR(Ours) & - & \textbf{88.7} & \textbf{94.4} \\ 
    \bottomrule 
  \end{tabular}
  \label{DukeMTMC-reID SOTA}
\end{center}
\end{table}

\textbf{DukeMTMC-reID.} DukeMTMC-reID is notably more challenging compared to Market-1501, with prevalent instances of occlusions in this dataset \cite{ristani2016performance}. As shown in Table. \ref{DukeMTMC-reID SOTA}, our method has surpassed all previously published results on the DukeMTMC-reID dataset, achieving an mAP and rank-1 accuracy of $88.7\%$ and $94.4\%$, respectively. Our performance exhibits a substantial lead over the state-of-the-art method, HLGAT \cite{zhang2021person}, with significant mAP and rank-1 enhancements of $+1.4\%$ and $+1.7\%$, respectively.

\begin{figure*}[ht]
\centering
\includegraphics[width=\linewidth]{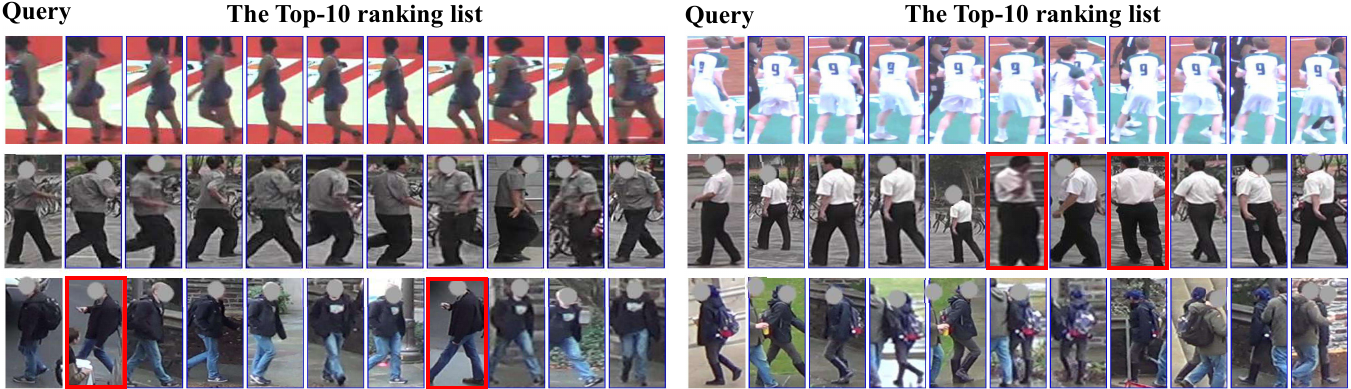}
\caption{The visualization of retrieval results for some samples from the candidate datasets. For each group, the first image represents the query, followed by the top-10 retrieval results from the gallery. Persons distinct from the query are highlighted with a red border. The first row of data is sourced from SynergyReID, the second row is from Market-1501, and the third row is from DukeMTMC-reID.}
\label{ranking list}
\end{figure*}

We can observe that the majority of previous state-of-the-art work has concentrated on extracting features from specific perspectives. For example, SOLIDER \cite{chen2023beyond} leverages prior semantic information, while HLGAT \cite{zhang2021person} extensively explores information between local features. In contrast, our approach extracts features from two distinct perspectives, facilitating adaptation to more complex scenarios. Furthermore, mutual distillation learning is employed in our model to enhance the feature representation capabilities under each viewpoint and improving the overall effectiveness of the model. To provide a more intuitive showcase of our model's effectiveness, we present some qualitative retrieval results from the test sets of candidate datasets in Figure. \ref{ranking list}. For each query, the top-10 retrieved images from the gallery are displayed in order, with those of non-matching identities outlined in red boxes for easy visualization. The first row contains samples from the SynergyReID dataset. Our model demonstrates strong robustness against variations in poses and gaits of the depicted individuals. In the second row with Market-1501 images, the model performs well on queries that are not strictly aligned. The third row shows examples on the DukeMTMC-reID dataset, where the second query in this row indicates the model can generalize well despite heavy occlusions.

\subsection{Ablation Study} \label{ablation}

\begin{figure}[ht]
\centering
\includegraphics[width=\linewidth]{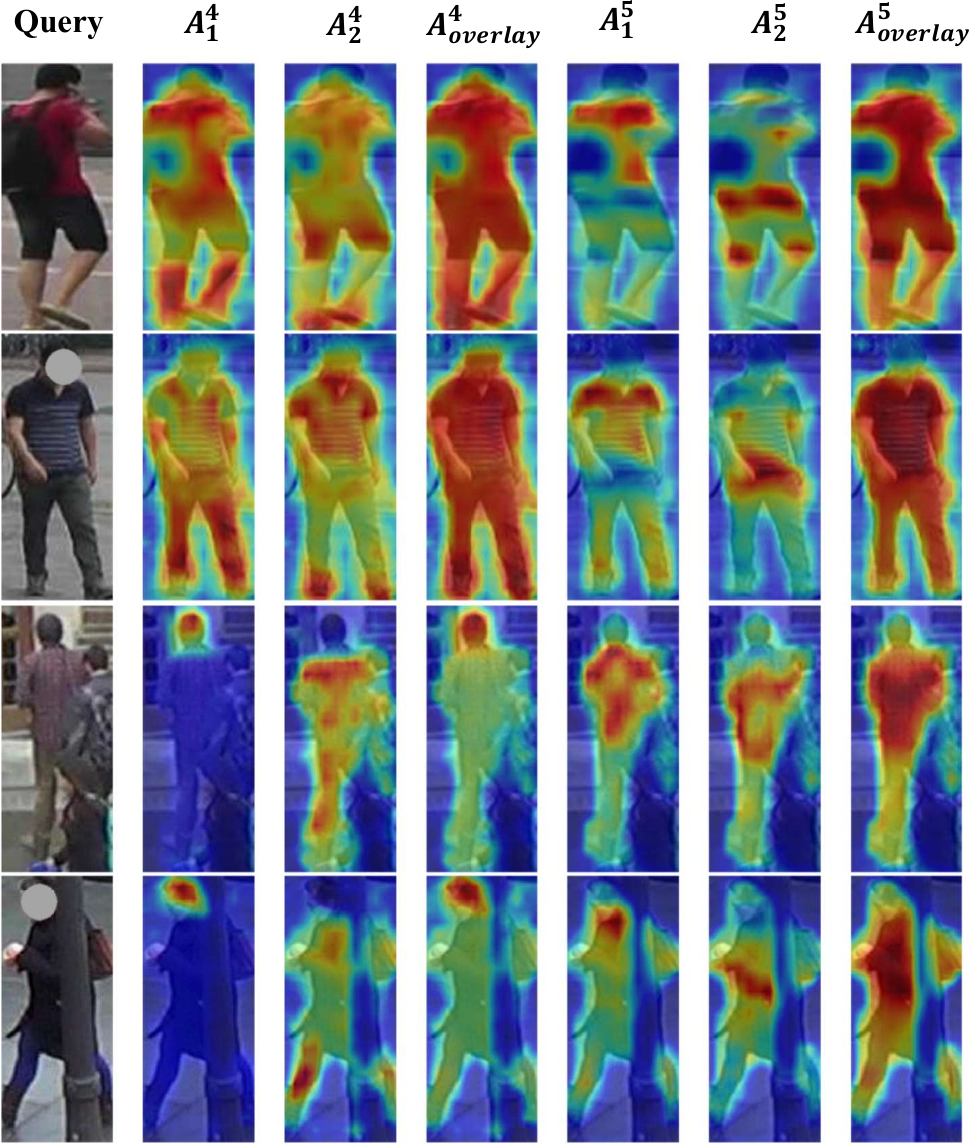}
\caption{The visualization of Soft Content Branch's attention maps $A^4$ and $A^5$. Red indicates high response values, while blue indicates low response values. $A^4_i$ represents the $i$-th attention map in $A^4$, while $A^4_{overlay}$ denotes the overlay of all heatmaps in $A^4$. The same notation applies to $A^5$. The data in the first two rows is sourced from Market-1501, while the data in the last two rows is from DukeMTMC-reID.}
\label{attn num}
\vspace{-3mm}
\end{figure}

\begin{table}
\begin{center}
  \caption{The performance of different embedding sizes on Market-1501 and DukeMTMC-reID datasets. $M$ refers to the embedding size.}
  \begin{tabular}{cccccc}
    \toprule
    \multirow{2}{*}{$M$} & \multicolumn{2}{c}{Market-1501} & \multicolumn{2}{c}{DukeMTMC-reID} \\
    & mAP & Rank-1 & mAP & Rank-1 \\
    \midrule
    256 & 93.7& 97.1& 87.8& 93.6\\
    512 & 94.2& \textbf{97.4}& \textbf{88.7}& \textbf{94.4}\\
    768 & \textbf{94.3}& 97.3& 88.5& 94.1\\
    \bottomrule
  \end{tabular}
  \label{embedding sizes}
\end{center}
\end{table}

\begin{table}
\begin{center}
  \caption{The performance impact of the different attention numbers on Market-1501 and DukeMTMC-reID datasets. $K$ refers to the number of attention maps.}
  \begin{tabular}{cccccc}
    \toprule
    \multirow{2}{*}{$K$} & \multicolumn{2}{c}{Market-1501} & \multicolumn{2}{c}{DukeMTMC-reID} \\
    & mAP & Rank-1 & mAP & Rank-1 \\
    \midrule
    2 & \textbf{94.2}& \textbf{97.4}& \textbf{88.7}& \textbf{94.4}\\
    4 & 94.1& 97.2& 88.5& 94.3\\
    8 & 94.0& 97.2& 88.4& 94.2\\
    \bottomrule
  \end{tabular}
  \label{Attention Num}
\end{center}
\end{table}

In order to delve deeper into the effectiveness of hyper-parameter settings and each module, a series of comprehensive ablation experiments is conducted on Market-1501 and DukeMTMC-reID datasets.

\subsubsection{Attention Number in Soft Content Branch}

The number of attention maps utilized by the Soft Content Branch is a crucial hyper-parameter in our model. These generated attention maps play a decisive role in distinguishing between foreground and background and focusing on key regions of the human body. To comprehensively understand the impact of the number of attention maps on our model's performance, we conduct an ablation study. Specifically, we vary the number of attention maps from 2, 4, to 8 and evaluate their effects on the model's accuracy. As shown in Table. \ref{Attention Num}, the model achieves its best performance when the number of attention maps is set to 2 for both the Market-1501 and DukeMTMC-reID datasets. Increasing the number of attention maps results in a slight decrease in model performance for both datasets.

To further illustrate how attention maps influence the model's ability to discern foreground and background, as well as concentrate on crucial regions, we visualize the attention maps in Figure. \ref{attn num}. As evident, for attention maps generated for an image, each of them does not encompass specific semantic information. However, the overlay of attention maps for both coarse-grained and fine-grained features, as indicated by $A^4_{overlay}$ and $A^5_{overlay}$ in Figure. \ref{attn num}, distinctly emphasizes the whole body of the target person in the image. Notably, different attention maps for one feature emphasize distinct regions corresponding to different body parts, highlighting the efficiency of our attention map generation. The first two rows of images in Figure. \ref{attn num} are derived from the Market-1501 dataset, showcasing our model's effective attention to individuals, whether from the front or the back. The latter two rows of images are from the DukeMTMC-reID dataset. Given the common occurrence of occlusion in DukeMTMC-reID, we specifically choose samples with occlusion-related challenges. The occlusions in the third row involve other pedestrians, while those in the fourth row are attributed to objects. When dealing with occlusion scenarios, our attention maps prove instrumental in highlighting the unobstructed parts of individuals, going beyond a mere focus on human bodies within the images. This capability allows us to effectively discriminate the target individual while minimizing the impact of non-target elements.

\begin{figure*}[ht]
\centering
\includegraphics[width=\linewidth]{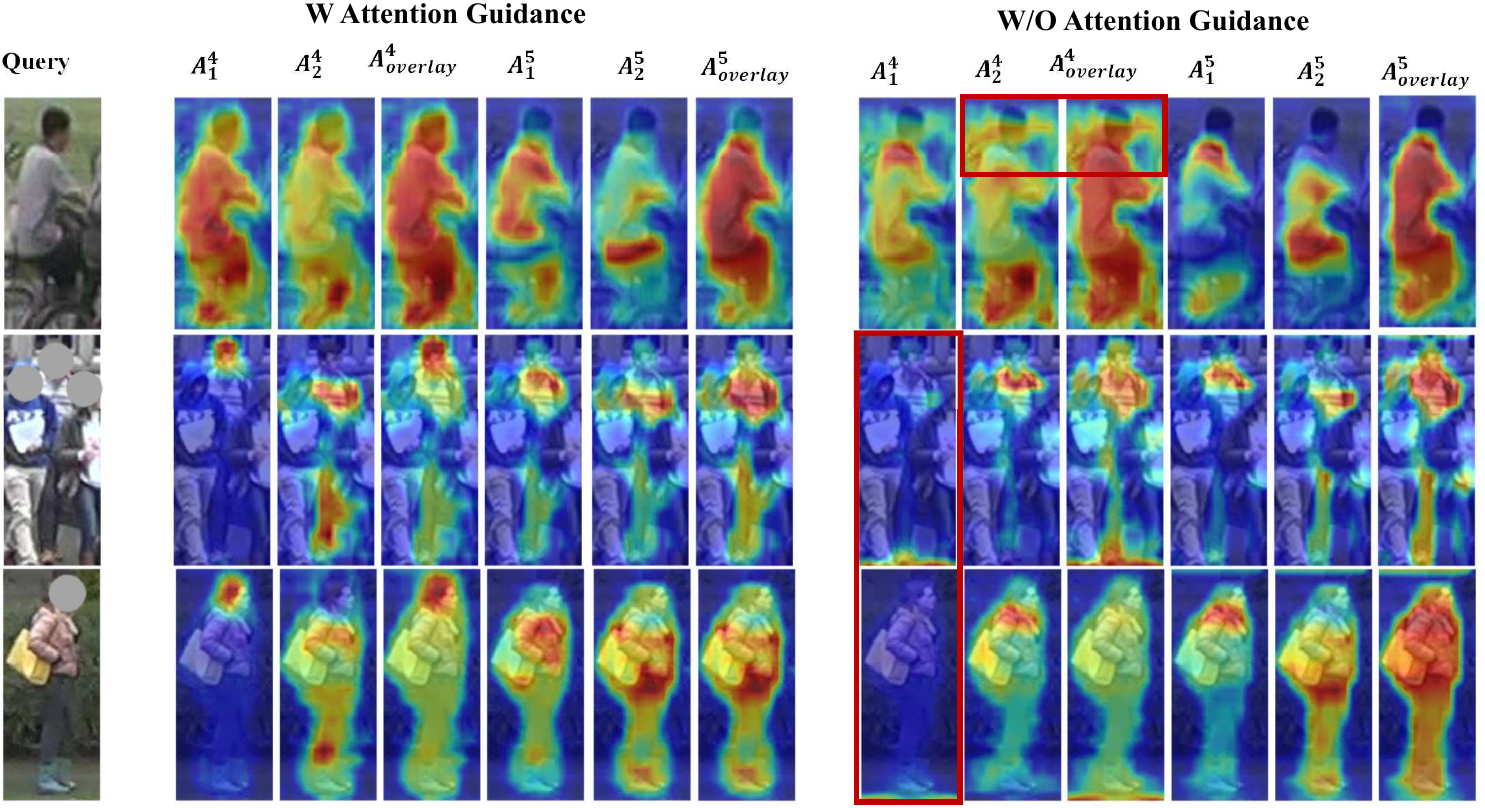}
\caption{The visualization of the attention maps $A^4$ and $A^5$ in the Soft Content Branch with and without the use of attention guidance. The first query image is sourced from Market-1501, while the subsequent two are from DukeMTMC-reID. Heatmaps, which notably lack emphasis on the human body or exhibit ineffectiveness, are specifically outlined with red bounding boxes. }
\label{attn guidance}
\end{figure*}

\begin{table}
\begin{center}
  \caption{The performance impact of attention guidance on Market-1501 and DukeMTMC-reID datasets.}
  \begin{tabular}{cccccc}
    \toprule
    \multirow{2}{*}{Attention Guidance} & \multicolumn{2}{c}{Market-1501} & \multicolumn{2}{c}{DukeMTMC-reID} \\
    & mAP & Rank-1 & mAP & Rank-1 \\
    \midrule
    w & \textbf{94.2}& \textbf{97.4}& \textbf{88.7}& \textbf{94.4}\\
    w/o & 94.1& 97.3& 88.4& 93.7\\
    \bottomrule
  \end{tabular}
  \label{attn guidance performance}
\end{center}
\end{table}

\begin{table}
\begin{center}
  \caption{The performance of individual feature and their combinations on Market-1501 and DukeMTMC-reID datasets. $f^{Hard}$ is the concatenation of features generated by the Hard Content Branch for reference, and similarly, $f^{Soft}$ represents the features from the Soft Content Branch. "Single" refers to a single network with the same setting as the branch in our model. }
  \begin{tabular}{cccccc}
    \toprule
    \multirow{2}{*}{Feature} & \multicolumn{2}{c}{Market-1501} & \multicolumn{2}{c}{DukeMTMC-reID} \\
    & mAP & Rank-1 & mAP & Rank-1 \\
    \midrule
    $f^{Hard}_{g^5}$       & 93.3& 97.0& 87.2& 93.7\\
    $f^{Hard}_{p^1}$       & 85.2& 95.5& 81.0& 90.8\\
    $f^{Hard}_{p^2}$       & 86.3& 93.6& 77.7& 88.4\\
    $f^{Hard}$             & 93.6& 97.2& 87.5& 94.1\\
    $f^{Hard}$ (Single)    & 93.2& 96.9& 87.1& 93.6\\
    \midrule
    $f^{Soft}_{g^5}$       & 93.1& 97.1& 86.4& 93.3\\
    $f^{Soft}_{g^4}$       & 93.0& 96.8& 86.5& 92.8\\
    $f^{Soft}$             & 93.6& 97.2& 87.1& 92.8\\
    $f^{Soft}$ (Single)    & 93.1& 96.6& 86.6& 93.0\\
    \midrule
    $f_{fusion}$           & 94.0& \textbf{97.4}& 88.4& 94.3\\
    \midrule
    $f^{Hard}$, $f^{Soft}$ & \textbf{94.2}& 97.3& 88.6& 94.3\\
    $f^{Hard}$, $f^{Soft}$, $f_{fusion}$ & \textbf{94.2} & \textbf{97.4} & \textbf{88.7}& \textbf{94.4}\\
    \bottomrule
  \end{tabular}
  \label{single feature performance}
  \vspace{-3mm}
\end{center}
\end{table}

\begin{table*}
\begin{center}
  \caption{The performance impact of Mutual Distillation Loss and Fusion Module on Market-1501 and DukeMTMC-reID datasets. "Individual Branch Networks" refers to two networks with the same setting as the Hard Content Branch and the Soft Content Branch in our model.}
  \begin{tabular}{cccccccc}
    \toprule
    \multirow{2}{*}{Model} & \multirow{2}{*}{Mutual Distillation Loss}  & \multirow{2}{*}{Fusion} & \multicolumn{2}{c}{Market-1501} & \multicolumn{2}{c}{DukeMTMC-reID} \\
    & & & mAP & Rank-1 & mAP & Rank-1 \\
    \midrule
    \multirow{4}{*}{MDPR(Ours)}   &  &  &  93.8& 97.1& 88.2& 94.0\\
                             & \checkmark &  &  93.9& 97.3& 88.1& 94.0\\
                             &  & \checkmark &  94.1& 97.2& 88.4& 93.8\\
                             & \checkmark & \checkmark  &  \textbf{94.2}& \textbf{97.4}& \textbf{88.7}& \textbf{94.4}\\
    \midrule
    \multirow{2}{*}{Individual Branch Networks} &  &  &  93.8& 97.1& 88.0& 93.9\\
    & \checkmark & \checkmark  &  94.1& 97.3& 88.5& 94.1\\
    \bottomrule
  \end{tabular}
  \label{distillation performance}
  \vspace{-3mm}
\end{center}
\end{table*}

\begin{table}
\begin{center}
  \caption{The performance impact of $\beta$ in Attention Diversity loss on Market-1501 and DukeMTMC-reID datasets.}
  \begin{tabular}{cccccc}
    \toprule
    \multirow{2}{*}{$\beta$} & \multicolumn{2}{c}{Market-1501} & \multicolumn{2}{c}{DukeMTMC-reID} \\
    & mAP & Rank-1 & mAP & Rank-1 \\
    \midrule
    0 & 94.1& 97.0& 88.6& 94.1\\
    0.1 & 94.1& \textbf{97.5}& 88.4& 94.0\\
    0.01 & 94.1& 97.1& 88.6& 94.2\\
    0.001 & \textbf{94.2}& 97.4& \textbf{88.7}& \textbf{94.4}\\
    0.0001 & 94.1& 97.1& 88.6& 94.2\\
    \bottomrule
  \end{tabular}
  \label{beta}
  \vspace{-3mm}
\end{center}
\end{table}

\subsubsection{Attention Guidance in Soft Content Branch}
In the Soft Content Branch, we employ an attention guidance mechanism to facilitate the generation of improved attention maps for coarse-grained features. More specifically, we use $A^5$ to guide the generation of $A^4$. To comprehensively investigate the impact of attention guidance, we assess its effects on performance in the context of deciding whether to employ attention guidance. As presented in Table. \ref{attn guidance performance}, for the Market-1501 dataset, the adoption of attention guidance leads to slight performance improvements, while for the DukeMTMC-reID dataset, employing attention guidance results in enhancements in mAP and Rank-1 accuracy, with improvements of $0.3\%$ and $0.7\%$, respectively.

Furthermore, we showcase the heatmap of attention maps with and without the application of attention guidance in Figure. \ref{attn guidance}. The first query image is from the Market-1501 dataset, while the latter two are from the DukeMTMC-reID dataset. It is evident that the utilization of the attention guidance mechanism significantly assists the model in directing its focus towards human body regions. For the first image of the query in Figure. \ref{attn guidance}, When attention guidance is absent, the attention maps tend to concentrate on areas near the target individual's neck, deviating from the regions containing the human body. For the latter two images, where attention guidance is not utilized, some of the generated attention maps, such as the attention map $A^4_1$, may be invalid. With the introduction of attention guidance, the generated attention maps become more accurate, making a crucial contribution to achieving improved performance.

\subsubsection{Hyper-Parameters in Attention Diversity Loss}
In the Attention Diversity loss, we introduce a regularization term to reduce the sensitivity of generated attention to noise and prevent overfitting. To determine the most suitable parameters, we conducted relevant ablation experiments for hyper-parameter $\beta$ and the experimental results are depicted in Table. \ref{beta}. For the Market-1501 dataset, Rank-1 is not sensitive to $\beta$, whereas mAP achieves high levels at 0.1 and 0.001. Conversely, for the DukeMTMC dataset, both mAP and Rank-1 reach their highest values when $\beta$ is set to 0.001. Taking all factors into consideration, we opt to set $\beta$ to 0.001.

\subsubsection{Embedding Size} 
The choice of embedding size $M$ has a direct impact on the final output dimension of the model, as evident in Table. \ref{feature dimension}. To comprehend the influence of varying embedding sizes on our model's performance, a comprehensive investigation is conducted. Specifically, we explore embedding sizes of 256, 512, and 768, assessing their effects on mAP and Rank-1. 

As depicted in Table. \ref{embedding sizes}, for the Market-1501 dataset, setting the embedding size to either 512 or 768 results in very similar outcomes. However, choosing a smaller embedding size leads to a reduction in model parameters and faster inference speed. Therefore, it is recommended to set the embedding size to 512. In contrast, for the DukeMTMC-reID dataset, an embedding size of 512 stands out as it achieves the highest values for both mAP and Rank-1 simultaneously.

\subsubsection{The Performance of Individual Features and Their Combinations}
To explore the contribution of single features and their combinations to the final output, we train the entire network and then conduct inference based on individual features and their combinations. Furthermore, to assess the impact of our approach on each individual branch, we train separate networks with settings identical to those of the individual branches in our model. The results are shown in Table. \ref{single feature performance}. It can be observed that the performance of both the Hard Content Branch and the Soft Content Branch on Market-1501 is similar, while on DukeMTMC-reID, the Hard Content Branch slightly outperforms the Soft Content Branch. Compared to using single networks with the same setting as the branches in our model, the branches in our model generally lead to superior performance for individual network. This can be attributed to the application of Knowledge Distillation and Fusion Module, which enhances the feature extraction capabilities of both branches. Additionally, on both datasets, ${f}_{fusion}$ surpasses both the Soft Content Branch and the Hard Content Branch, closely approaching the results obtained when simultaneously using $f^{Hard}$ and $f^{Soft}$.

\subsubsection{The Effectiveness of Knowledge Distillation and Fusion Module}
To further investigate the impact of the Mutual Distillation Loss and Fusion Module, ablation experiments are conducted based on our model and the scenarios of two separate branch networks. The experimental results are presented in the Table. \ref{distillation performance}. For our model, using only Mutual Distillation Loss or Fusion Module alone does not necessarily lead to performance improvement. However, combining both yields superior results for the two datasets. Moreover, the simultaneous use of both also enhances the performance of Individual Branch Networks.

\section{Conclusion}
In conclusion, we propose a dual-branch framework for person ReID that integrates multiple perspectives within a unified model. The Hard Content Branch employs a uniform horizontal partitioning technique to capture local features, while the Soft Content Branch incorporates a carefully designed attention mechanism, enhancing foreground-background differentiation and facilitating the extraction of crucial global features across multiple granularities. Leveraging the Knowledge Mutual Distillation and Fusion Module, our framework enables knowledge sharing among branches, and fuses the outputs from each branch to obtain a more comprehensive feature representation. Our approach is evaluated on widely used person ReID datasets, establishing its effectiveness and superiority, with performance metrics reaching or surpassing the state-of-the-art methods.
\ifCLASSOPTIONcaptionsoff
  \newpage
\fi

{
    \small
    \bibliographystyle{IEEEtran}
    \bibliography{IEEEabrv, ref}
}




\end{document}